\documentclass[conference]{IEEEtran}
\IEEEoverridecommandlockouts

\usepackage{url}            
\usepackage{xcolor}         
\usepackage{graphicx}
\usepackage{wrapfig}
\usepackage{amsmath,bm}
\usepackage{enumerate}
\usepackage{bm}
\usepackage{algorithm}
\usepackage{algorithmic}
\usepackage{array}
\usepackage{amsthm}
\usepackage{scalefnt}
\usepackage{subfigure}
\usepackage{makecell}
\usepackage{mathrsfs}
\usepackage{wrapfig, flushend, multirow}
\usepackage{lipsum}
\usepackage{enumitem,calc}
\usepackage{cite}
\usepackage[numbers,sort&compress]{natbib}
\usepackage{caption}
\usepackage{amsmath,amssymb,amsfonts}
\usepackage{textcomp}
\usepackage{perpage}
\MakePerPage{footnote}

\usepackage{hhline}
\def\BibTeX{{\rm B\kern-.05em{\sc i\kern-.025em b}\kern-.08em
    T\kern-.1667em\lower.7ex\hbox{E}\kern-.125emX}}
    
\newif\ifshowcomment
\showcommenttrue

\newtheorem{problem}{Problem}

\makeatletter

\newcommand\preitem{\mdseries\textbullet\space}
\usepackage[input-decimal-markers={,},round-mode=places,round-precision=3]{siunitx}
\newlist{desclist}{description}{3}
\setlist[desclist,1]{format=\preitem\bfseries,leftmargin=\widthof{\preitem},style=sameline}

\usepackage{expl3}
\ExplSyntaxOn
\newcommand\latinabbrev[1]{
  \peek_meaning:NTF . {
    #1\@}%
  { \peek_catcode:NTF a {
      #1.\@ }%
    {#1.\@}}}
\ExplSyntaxOff

\def\eg{\latinabbrev{e.g}}

\def\ie{\latinabbrev{i.e}}
\newcommand{\name}{{\textsc{LogiRec}}}

\usepackage[belowskip=-10pt,aboveskip=-0pt]{caption}
\begin{document}

\title{Towards High-Order Complementary Recommendation via Logical Reasoning Network}

\author{
\IEEEauthorblockN{1\textsuperscript{st} Given Name Surname}
\IEEEauthorblockA{\textit{dept. name of organization (of Aff.)} \\
\textit{name of organization (of Aff.)}\\
City, Country \\
email address or ORCID}
}

\author{\IEEEauthorblockN{Longfeng Wu}
\IEEEauthorblockA{
\textit{Virginia Tech}\\
Blacksburg, VA, United States \\
longfengwu@vt.edu}
\and
\IEEEauthorblockN{Yao Zhou}
\IEEEauthorblockA{\textit{Instacart Inc.} \\
San Francisco, CA, United States \\
yaozhou3@illinois.edu}
\and
\IEEEauthorblockN{Dawei Zhou}
\IEEEauthorblockA{
\textit{Virginia Tech}\\
Blacksburg, VA, United States \\
zhoud@vt.edu}
}

\maketitle

\begin{abstract}
Complementary recommendation gains increasing attention in e-commerce since it expedites the process of finding frequently-bought-with products for users in their shopping journey. Therefore, learning the product representation that can reflect this complementary relationship plays a central role in modern recommender systems. In this work, we propose a logical reasoning network, \name, to effectively learn embeddings of products as well as various transformations (projection, intersection, negation) between them. 
\name\ is capable of capturing the asymmetric complementary relationship between products and seamlessly extending to high-order recommendations where more comprehensive and meaningful complementary relationship is learned for a query set of products. Finally, we further propose a hybrid network that is jointly optimized for learning a more generic product representation. We demonstrate the effectiveness of our \name\ on multiple public real-world datasets in terms of various ranking-based metrics under both low-order and high-order recommendation scenarios. 
\end{abstract}


\section{Introduction}
\begin{figure}[t] 
   \captionsetup{font=footnotesize}
    \centering
    \includegraphics[width=\linewidth]{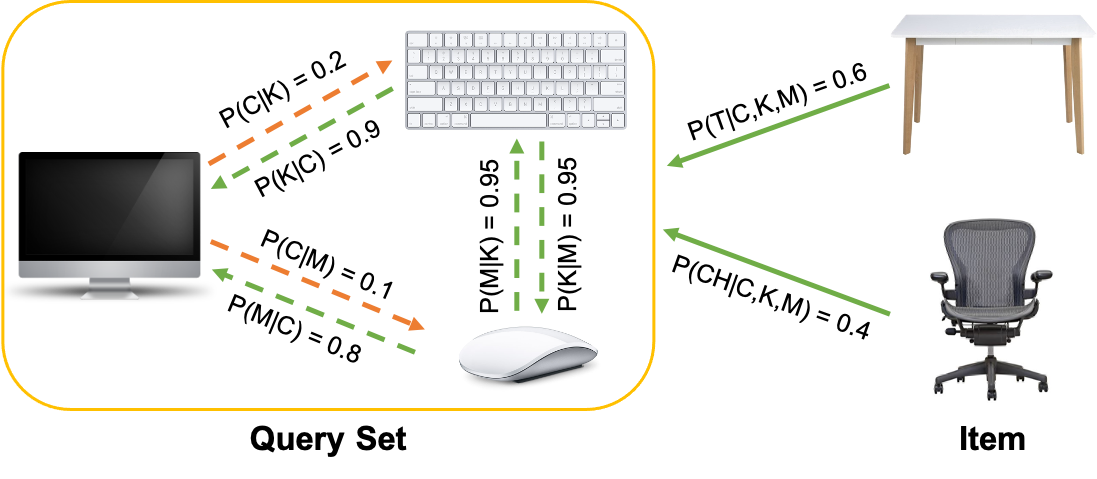}
    \caption{An illustrative example of complementary recommendation. Given a set of query items {\tt computer(C)}, {\tt keyboard(K)}, {\tt mouse(M)} with low-order complementary relationship (dashed arrow) among them, the system recommends a product {\tt table(T)} with the high score {\tt P(T|C,K,M)} that is a good high-order complement (solid arrow) to the query set over another product {\tt chair(CH)}.}
    \label{fig:example}
    \vspace{-2mm}
\end{figure}

Driven by the rapid growth of e-commerce business, recommender systems (RS) have become an indispensable component of our modern life. Recommending relevant products to customers by learning from their personalized content has been well studied in recent years with solid progress. Nevertheless, most e-commerce platforms offer a wide range of products and their selling curves are always long-tailed due to the effect of position bias (e.g., customers always favor the high-ranked popular products). Therefore, researchers and practitioners start resorting to complementary recommendation (CR) for product exploration and increasing cross-selling between different categories. Complementary recommendation mainly relies on learning the underlying relationship among products, which is often characterized by the co-purchase and co-view patterns from customer engagement traffic. To extract the complementary relationship between products, there exists a few prior works such as association mining~\cite{wu2007association}, item-based collaborative filtering~\cite{sarwar2001item}, representation learning~\cite{kang2019complete, xu2020knowledge,hao2020p,zhou2021pure}. However, the traditional data-mining and collaborative filtering based approaches~\cite{wu2007association, sarwar2001item} can only learn the symmetrical co-purchase relationship between products; most existing deep models rely on auxiliary information (e.g., scene background features required in~\cite{kang2019complete}, user purchase sequence information required in~\cite{xu2020knowledge}, product catalog information~\cite{hao2020p}) which limits their models' adoption to a wider application domain. 
Moreover, the majority of them can only infer the low-order complementary relationship between products but cannot learn a query set's high-order complementary relationship. For example, in Figure~\ref{fig:example}, a customer who purchases a computer is highly like to purchase a keyboard as well but not vice versa, i.e., P(keyboard$|$computer) $>$ P(computer$|$keyboard). From another perspective, given a customer who purchased a computer, mouse, and keyboard, it is highly likely he/she is planning on building a home office, therefore P(table$|$(computer, keyboard, mouse)) should be high in order to capture this high-order complementary relationship. 

To find answers to the aforementioned challenges, we utilize several first-order logical reasoning operations and learn probabilistic embedding of products. These logical operations are designed to be learnable in the embedding space over both entities \& query sets and are capable of satisfying these appealing properties: \textbf{Asymmetry}: if we utilize $\rightarrow$ as the notation for `complementary' relationship, then, asymmetry means for two products $A$ and $B$, $A \rightarrow B$ does not imply $B \rightarrow A$; \textbf{High-order}: the complementary products for a query set of product combos are supposed to be more comprehensive and meaningful compared with each individual product's complements. 

Inspired by recent advances in first-order logical reasoning in~\cite{liu2021neural,ren2020beta}, in this work, we propose a novel framework, \name, which embeds each product as a vector with beta distribution, and leverages three basic logical operations (projection, intersection, negation) to demonstrate the transformation between products. 
Different from BetaE, logical operations here are utilized to improve the quality of the complementary relationship between products.
For low-order CR, we leverage the combination of three basic logical operations to infer the complementary items of the query item. For high-order CR, we develop an attention mechanism to summarize the logical reasoning based on each item in the query set and then conduct complementary projection via a transformation neural network. On top of that, a hybrid model is built to capture both types of complementary relationships. 
The main contributions of this paper are summarized as follows:
\begin{itemize}
    \item {\bfseries Problem:} We formalize the \emph{high-order complementary recommendation} problem and identify the unique challenges inspired by the real applications.  
    \item {\bfseries Model:} We propose an end-to-end framework \name, which jointly models the asymmetric nature and high-order dependencies between the query set and its potential complementary items.
    \item {\bfseries Evaluation:} We systematically evaluate the performance of \name\ on two settings: 1) lower-order CR and 2) high-order CR. Extensive results prove the superior performance of \name\ under both scenarios.
\end{itemize}

The rest of this paper is organized as follows. We provide the notation and problem definition in Section 2, before introducing the proposed framework in Section 3. The experimental setup and results are discussed in Section 4, followed by the literature review in Section 5. Finally, we conclude this paper in Section 6.

\section{Problem Definition}

\textbf{Notations} We use upper case calligraphic font letters to denote graphs (e.g. $\mathcal{G}$), bold lower case letters to denote vectors (e.g. $\bm{\theta}$), and regular lower case letters to denote scalars (e.g. $\gamma$). A product graph is denoted as $\mathcal{G} = (\mathcal{I}, \mathcal{E})$, where $\mathcal{I}$ is the node set of items, and $\mathcal{E} = \{\mathcal{E}_{cp}, \mathcal{E}_{cv}\}$ is the edge set representing two types of interactions between items (\ie, co-purchase $\mathcal{E}_{cp}$ and co-view $\mathcal{E}_{cv}$). 
$\mathcal{Q}$ is the query set, and $\overline{\mathcal{Q}}$ is the complementary item set of $\mathcal{Q}$.

\textbf{Problem Definition} Given a query item $q \in \mathcal{I}$, existing CR systems~\cite{kang2019complete,hao2020p,xu2020knowledge} aim to find a collection of candidate items $\overline{\mathcal{Q}}\subseteq \mathcal{I}$ that well complement the query item $q$. If we consider each item in $\mathcal{G}$ as a distinct state,  we can interpret the complementary recommendation problem as the $1^\text{st}$-order Markov chain that aims to estimate the conditional probability $P(S_{t+1} = i | S_t = q)$ over all potential items $\{i| i \in \mathcal{I}, i \neq q\}$, where both $S_t$ and $S_{t+1}$ denote the hidden states.

However, it is often the case that the recommendation systems may obtain a set of queries from customers, which provides more context (\ie, the relationship and dependencies between items in the query set) for us to understand the customers' real intentions and thus enables us to make more accurate recommendations. 
Motivated by this, we generalize the problem to the high-order setting, where the query is represented as a set of items. 
To be specific, given a query set $\mathcal{Q} = \{q_1, q_2, \ldots, q_{|\mathcal{Q}|}\}$ with $|\mathcal{Q}|$ items, we propose to develop a CR system that accurately predicts the potential items with high $|\mathcal{Q}|^{\text{th}}$-order conditional probability~\cite{DBLP:conf/kdd/ZhouZYATDH17,DBLP:conf/kdd/FuZH20}, as shown in Equation~\ref{eqn:conditional_pro}.
\begin{equation}
\small
\label{eqn:conditional_pro}
\begin{split}
P(S_{t+1} = i | S_t = q_1, S_{t-1} = q_2, \ldots, S_{t-|\mathcal{Q}| +1} = q_{|\mathcal{Q}|})
\end{split}
\end{equation}
We refer to this problem as high-order complementary recommendation (HCR), which is formally defined in Problem 1. 
It is worthy to mention that conventional CR could be considered as a special case of HCR when the query set $\mathcal{Q}$ includes only one item, \ie, $|\mathcal{Q}| = 1$.

{\setlength{\parindent}{0pt}
\begin{problem}
 \textbf{High-Order Complementary Recommendation} \\
    \textbf{Input:} (i) a product graph $\mathcal{G} = (\mathcal{I}, \mathcal{E})$,  (ii) a query set $\mathcal{Q} = \{q_1, \ldots, q_{|\mathcal{Q}|}\}$, (iii) the number of desired recommendations $|\overline{\mathcal{Q}}|$.\\
    \textbf{Output:} the set of complementary items $\overline{\mathcal{Q}}$.
\end{problem}
}

\section{Model}
In this section, we present our proposed \name\ framework for the complementary recommendation. The core idea of \name\ is to learn the high-order complementary transformation by leveraging three basic logical operations (projection, intersection, and negation) upon probabilistic representations of products. 

\subsection{A Generic Learning Framework}
\begin{figure}[t]
  \captionsetup{font=footnotesize}
   \centering
   \includegraphics[width=\linewidth]{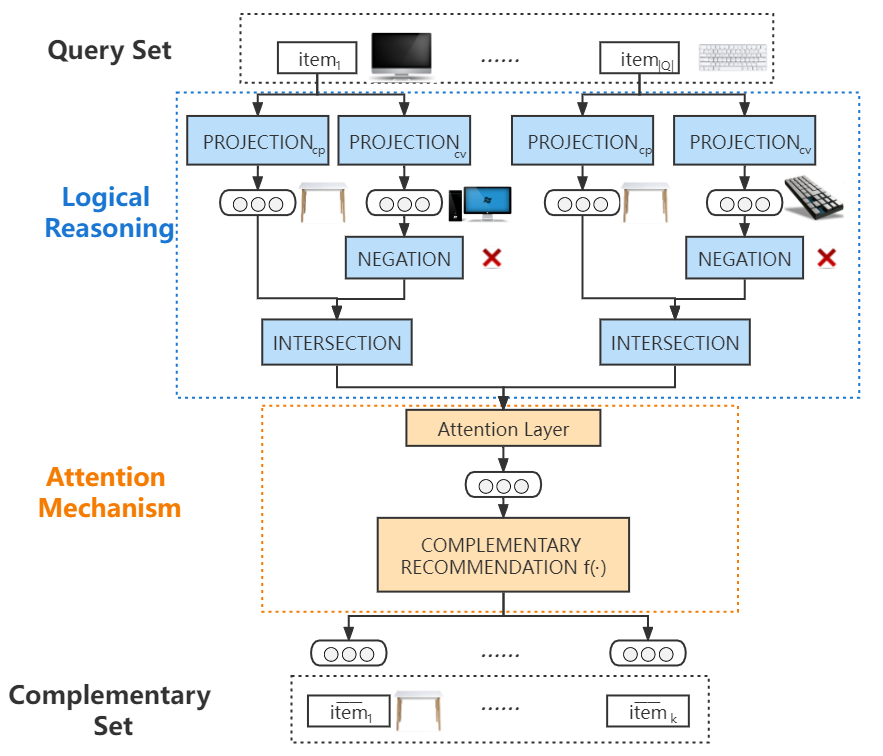}
   \caption{An illustration of the proposed framework \name, which is composed of a logical reasoning module (M1) for asymmetric recommendations and an attention mechanism (M2) for the high-order CR.}
   \label{fig:framework}
 \end{figure}

An overview of our framework \name\ is shown in Figure~\ref{fig:framework}. Essentially, the design of \name\ aims to provide an end-to-end solution to address Problem 1 by jointly leveraging the asymmetric nature and high-order dependencies between the query set $\mathcal{Q}$ and the potential complements $\overline{\mathcal{Q}}$. 
In particular, to accommodate the asymmetry property of complementary recommendation, we propose the logical reasoning module in M1 to automatically filter out item pairs that exhibit substitute relationships from co-purchase data via neural logical operations. To understand the high-order complementary transformation among items, we develop the attention mechanism in M2 to learn the high-order dependencies between query sets and the potential complements. 

\textbf{M1. Logical Reasoning for Asymmetric Recommendations. }
Recommendation systems aim to learn a mapping function $f(\cdot)$ that projects the query item $q$ to the target item $i$. Despite the long-standing research in recommendation systems, many approaches~\cite{mcauley2015inferring,he2020lightgcn} are based on symmetric measures (\eg, cosine similarity and Euclidean distance), thus fail to accommodate the asymmetric nature of CR (\eg, $P(C | K) \neq P(K | C)$ in Figure~\ref{fig:example}). Here, we model the asymmetric relationships in complementary recommendation from a new perspective, \ie, \emph{logical reasoning}. 
In this paper, we propose to replace conventional symmetric measures with learnable logical operators. 
The logical operators (except negation) defined here are directed and irreversible, so our method can accommodate the asymmetric nature of CR.

Moreover, in modern e-commerce networks, it is often the case that items exhibit rich semantic meanings, which results in different levels of uncertainties. For example, apple can refers to both a kind of fruit and an electronic brand, while banana is less likely to cause confusion, so the certainty of these two items is different. Motivated by this observation,
we propose to represent items with probabilistic embeddings to model uncertainties and capture their rich semantic meanings. We formulate the embedding of each item $i \in \mathcal{I}$ as a collection of $d$ independent Beta distributions $\{Beta(\alpha_{i,1}, \beta_{i,1}), \ldots, Beta(\alpha_{i,d}, \beta_{i,d})\}$ with $2d$ learnable parameters. 
The key advantage of using Beta embedding is that Beta distribution supports closed-loop logical operations~\cite{ren2020beta}. 
For the simplicity of notation and presentation, we introduce this paper by denoting $Beta(\bm{\alpha}_i, \bm{\beta}_i) = \{Beta(\alpha_{i,1}, \beta_{i,1}), \ldots, Beta(\alpha_{i,d}, \beta_{i,d})\}$.
With that, we can formulate the mapping function $f(\cdot)$ between query item $q$ and complementary item $i$ with a transformation neural network, \eg, multi-layer perceptron (MLP). 
\begin{equation}
\label{eqn:mlp_cr}
\begin{split}
 f(\bm{\theta}_{q, Beta(\bm{\alpha}_q, \bm{\beta}_q)} )  & = \bm{\theta}_{i, Beta(\bm{\alpha}_i, \bm{\beta}_i)}
\end{split}
\end{equation}
where $\bm{\theta}_{q, Beta(\bm{\alpha}_q, \bm{\beta}_q)}$ and $\bm{\theta}_{i, Beta(\bm{\alpha}_i, \bm{\beta}_i)}$ are the Beta embedding of query item $q$ and the complementary item $i$. 

In practice, it is reasonable to assume that customers will view several similar items before making the final decision, so co-view data displays a more interchangeable relationship
between items~\cite{mcauley2015inferring}, while co-purchase relationships show more complementary properties of items since they are often purchased together to meet a specific need. To parameterize the mapping function $f(\cdot)$, one common strategy is to use item co-purchase records $\mathcal{E}_{cp}$ as the ground truth labels for CR. However, as being pointed out by~\cite{hao2020p,xian2021ex3}, there is often an overlap between co-purchase data and co-view data. For example, a customer may check two different shirts and purchase them together. Under such circumstances, the data may reveal that these two items have both the co-view and co-purchase relationships, which could add noise to the product graph and deteriorate the model's performance. That is being said, excluding the overlapped items $\mathcal{E}_{cp} - (\mathcal{E}_{cp} \cap \mathcal{E}_{cv})$ would potentially improve the quality of the ground truth labels for training $f(\cdot)$. The pruning process can be re-written as follows. 

\begin{equation} 
\small
\label{eq:pruning}
\begin{split}
\mathcal{E}_{cp} - (\mathcal{E}_{cp} \cap \mathcal{E}_{cv})
& = (\mathcal{E}_{cp} \cap \overline{\mathcal{E}_{cp}}) \cup (\mathcal{E}_{cp} \cap \overline{\mathcal{E}_{cv}}) 
= \mathcal{E}_{cp} \cap \overline{\mathcal{E}_{cv}}
\end{split}
\vspace{-1mm}
\end{equation}

Instead of manually removing the overlapped items, we propose to parameterize the above pruning process via learnable logical operators, and the mapping function $f(\cdot)$ can be learned in an end-to-end manner. Particularly, we define three neural logical operators in Equations~\ref{eqn:mlp},~\ref{eqn:neg},~\ref{eqn:inter}, which can be directly performed over the learned Beta embeddings of items. 

\begin{desclist}
    \item \textbf{Projection:} 
    Given a beta embedding $\bm{\theta}_{q, Beta(\bm{\alpha}_q, \bm{\beta}_q)}$ of the query item $q$ and a relationship $r$, the projection operator outputs a new beta embedding $\bm{\theta}_{i, Beta(\bm{\alpha}_{i}, \bm{\beta}_{i})}$ of item $i$, where items $q$ and $i$ are connected by relationship $r$. 
    \begin{equation} 
    \small
    \label{eqn:mlp}
    \begin{split}
    {\tt PROJECTION}_{r}(\bm{\theta}_{q, Beta(\bm{\alpha}_q, \bm{\beta}_q)}) &= \bm{\theta}_{i, Beta(\bm{\alpha}_{i}, \bm{\beta}_{i})} 
    \end{split}
    \end{equation}
where $r$ can be either co-view or co-purchase relationship.
      
    \item \textbf{Negation:} 
    Given a beta embedding $\bm{\theta}_{q, Beta(\bm{\alpha}_q, \bm{\beta}_q)}$, the negation operator outputs a new beta embedding $\bm{\theta}_{i, Beta(\bm{\alpha}_{i}, \bm{\beta}_{i})}$. 
    \begin{equation} 
    \small
    \label{eqn:neg}
    \begin{split}
    {\tt NEGATION}(\bm{\theta}_{q, Beta(\bm{\alpha}_{q}, \bm{\beta}_{q})}) = \bm{\theta}_{i, Beta(\bm{\alpha}_{i}, \bm{\beta}_{i})}
    \end{split}
    \end{equation}
    where $\bm{\alpha}_{q} = 1/\bm{\alpha}_{i}, \bm{\beta}_{q} = 1/\bm{\beta}_{i}$.
    
    \item \textbf{Intersection:} 
    Given a set of beta embedding query items  $\{\bm{\theta}_{q_{j}, Beta(\bm{\alpha}_{q_{j}}, \bm{\beta}_{q_{j}})} | j= 1, \ldots, n\}$, the intersection operator outputs a new beta embedding $\bm{\theta}_{i, Beta(\bm{\alpha}_{i}, \bm{\beta}_{i})}$, where $i$ is connected to every $q$ from the set.
    \begin{equation} 
    \small
     \label{eqn:inter}
    \begin{split}
    &{\tt INTERSECTION}(\{\bm{\theta}_{q_{j}, Beta(\bm{\alpha}_{q_{j}}, \bm{\beta}_{q_{j}})} | j= 1, \ldots, n\})
    = \bm{\theta}_{i, Beta(\bm{\alpha}_{i}, \bm{\beta}_{i})}
    \end{split}
    \end{equation}
    where $\bm{\alpha}_{i} = \sum_{j=1}^{n} w_{j}\bm{\alpha}_{q_{j}},
     \bm{\beta}_{i} = \sum_{j=1}^{n} w_{j}\bm{\beta}_{q_{j}}$,
    and aggregation weights $w_{j}$ are computed via softmax: 
     $w_{j} = \frac{exp(\bm{\theta}_{j, Beta(\bm{\alpha}_{j}, \bm{\beta}_{j})})} {\sum_{k=1}^{n} exp(\bm{\theta}_{k, Beta(\bm{\alpha}_{k}, \bm{\beta}_{k})}}$.\\
\end{desclist}

\textbf{M2. Attention Mechanism for High-Order Complementary Recommendation. } 
Recall that, given a query with a set of items $\mathcal{Q} = \{q_1, q_2, \ldots, q_{|\mathcal{Q}|}\}$, our goal is to identify the top complementary items that maximize the conditional probability $P(S_{t+1} = i | S_t = q_1, S_{t-1} = q_2, \ldots, S_{t-|\mathcal{Q}| +1} = q_{|\mathcal{Q}|})$. The key challenge is how to model high-order dependencies between the query set $\mathcal{Q}$ and the complementary set $\overline{\mathcal{Q}}$. Most of current recommendation systems~\cite{hao2020p, xu2020knowledge} deal with one query item, which is hard to capture the customers' intentions. The example in Figure~\ref{fig:example} indicates that a query set with multiply items could reveal more information than the items themselves.
As shown in the orange region of Figure~\ref{fig:framework}, here we develop an attention mechanism that automatically learns the weight of each item in $\mathcal{Q}$ and then summarizes them into a unique representation (\ie, Beta embedding) for the whole query set. 
In particular, the attention mechanism takes the Beta embedding of each query item in the query set as input and then outputs the corresponding attention weight $w_{j}$ as follows. 
\begin{equation} 
\small
\label{eq:low_order_attention}
\begin{split}
w_{j} = \frac{exp(\bm{\theta}_{j, Beta(\bm{\alpha}_{j}, \bm{\beta}_{j})})}{\sum_{k=1}^{|\mathcal{Q}|}exp(\bm{\theta}_{k, Beta(\bm{\alpha}_{k}, \bm{\beta}_{k})})}
\end{split}
\end{equation}
where $j=1, \ldots, |\mathcal{Q}|$. 
With the learned attention weights $w_{j}$ over all the query items, we can compute the summarization representation $\bm{\theta}_{\mathcal{Q}, Beta(\bm{\alpha}_\mathcal{Q},\bm{ \beta}_\mathcal{Q})}$ of the entire query set ${\mathcal{Q}}$,
where $\bm{\alpha}_{\mathcal{Q}} = \sum_{j=1}^{|\mathcal{Q}|} w_{j}\bm{\alpha}_{j},
 \bm{\beta}_{\mathcal{Q}} = \sum_{j=1}^{|\mathcal{Q}|}w_{j}\bm{\beta}_{j}$.

Note that $\bm{\theta}_{i, Beta(\bm{\alpha}_i, \bm{\beta}_i)}$ may not be associated with an exact item in the product graph $\mathcal{G}$. In our implementation, we retrieve the top-$|\overline{Q}|$ potential complementary items by searching the items with the smallest KL divergence to $\bm{\theta}_{i, Beta(\bm{\alpha}_i, \bm{\beta}_i)}$.

\subsection{Optimization algorithm}
The overall learning objective of \name\ is designed based on the negative sampling~\cite{mikolov2013distributed} as follows. 
\begin{equation} 
\label{eqn:loss}
\begin{split}
\mathcal{L} = &-log \sigma (\gamma - KL(\bm{\theta}_{i, Beta(\bm{\alpha}_{i}, \bm{\beta}_{i})};f(\bm{\theta}_{\mathcal{Q}, Beta(\bm{\alpha}_\mathcal{Q}, \bm{\beta}_\mathcal{Q})})) \\
&- \sum_{j=1}^{m}\frac{1}{m}log \sigma (KL(\bm{\theta}_{i', Beta(\bm{\alpha}_{i'}, \bm{\beta}_{i'})};f(\bm{\theta}_{\mathcal{Q}, Beta(\bm{\alpha}_\mathcal{Q}, \bm{\beta}_\mathcal{Q})}))- \gamma)
\end{split}
\end{equation}
where $KL$ computes the KL divergence between two distributions, $i$ represents the complementary item for query set $\mathcal{Q}$, $i' \in \mathcal{I}$ is a random negative sample that has nothing to do with the query set $\mathcal{Q}$, $m$ denotes the number of negative samples, and $\gamma$ denotes the decision margin. 
The first KL divergence models the distance between the query set and its answer, and the second one models the distance between the query set and the random items. Our objective is to minimize the first one while maximizing the second one. 
It is also worth mentioning that, different from cosine similarity and Euclidean distance, KL divergence is naturally an asymmetric measure, which guarantees the asymmetry property of \name.

\section{Experiments}
In this section, we compare {\name} with its variants and several popular baselines in terms of their low-order and high-order CR performances on six publicly accessible datasets.

\begin{table}[h]
\vspace{2.5mm}
    \setlength\tabcolsep{3.5pt} 
    \captionsetup{font=footnotesize}
     \caption{DATASET STATISTICS. The number of co-purchase pairs is denoted as $|\mathcal{E}_{cp}|$, the number of co-view pairs is denoted as $|\mathcal{E}_{cv}|$, density is computed as $(|\mathcal{E}_{cp}| + |\mathcal{E}_{cv}|)/|\mathcal{I}|^2$.} 
    \begin{center}
    \begin{tabular}{|c|c|c|c|c|c|}
        \hline
        \textbf{Dataset} & \# of items & $|\mathcal{E}_{cp}|$ & $|\mathcal{E}_{cv}|$ & $  |\mathcal{E}_{cp} - \mathcal{E}_{cv}|$ & density\\
        \hline \hline
        \textbf{Gift Cards} & 3.9K & 20.9K & 33.4K & 11.6K & $3.56 e^{-3}$\\
        \hline
        \textbf{Digital Music} & 91.4K & 94.6K & 98.1K& 72.6K & $2.31 e^{-5}$\\
        \hline
        \textbf{Video Games} & 116.4K & 773.5K & 763.5K& 605.0K & $1.13 e^{-4}$ \\
        \hline
        \textbf{Pet Supplies} & 324.5K & 981.3K & 2.03M& 780.1K & $2.86 e^{-5}$ \\
        \hline
        \textbf{Amazon Fashion} & 421.5K & 272.4K & 316.9K& 212.1K & $3.32 e^{-6}$ \\
        \hline
        \textbf{Office Products} & 657.8K & 2.42M & 1.85M& 2.01M & $9.87 e^{-6}$ \\
        \hline
    \end{tabular}
      \label{tab:dataset}
    \end{center}
    \vspace{-4mm}
\end{table}

\begin{table*}[!t]
\captionsetup{font=small}
    \caption{Evaluation results for low-order complementary recommendation. ({\tt OOME} stands for out-of-memory error)} \label{tab:low-order}
    \centering
    \begin{tabular}{|c|ccc|ccc|ccc|}
        \hline
        \textbf{Dataset} &
        \multicolumn{3}{c|}{\textbf{Gift Cards}} &
        \multicolumn{3}{c|}{\textbf{Digital Music}} &
        \multicolumn{3}{c|}{\textbf{Amazon Fashion}}  \\
        \hline
        \textbf{Metrics} & 
        Hit@3 & NDCG@3 & MRR &
        Hit@3 & NDCG@3 & MRR &
        Hit@3 & NDCG@3 & MRR \\
        \hline
        \textbf{CF} & 0.009 & 0.005 & 0.024& 0.096 & 0.061 & 0.074 & 0.062 & 0.035 & 0.032\\
        \textbf{MF} & 0.011 & 0.006 & 0.024& 0.075 & 0.060 & 0.064 & 0.016 & 0.012 & 0.013\\
        \textbf{Knowledge-Aware} & 0.009 & 0.004 & 0.025& 0.075 & 0.053 & 0.062 & 0.039 & 0.023 & 0.024\\
        \textbf{LightGCN} & 0.119 & 0.047 & 0.074& 0.222 & 0.213 & 0.222 & {\tt OOME} & {\tt OOME} & {\tt OOME}\\ \hline
        \textbf{\name$_{\mathrm{Low}}$} & 0.121 & 0.088 & 0.129& 0.550 & 0.469 & 0.479& 0.182 & 0.142 & 0.161\\
        \textbf{\name$_{\mathrm{High}}$} & 0.027 & 0.021 & 0.030 & 0.160 & 0.126& 0.134 & 0.116 & 0.096 & 0.103\\
        \textbf{\name$_{\mathrm{Hybrid}}$} & \textbf{0.155} & \textbf{0.110} & \textbf{0.149}& \textbf{0.567} & \textbf{0.484} & \textbf{0.493}& \textbf{0.278} & \textbf{0.226} & \textbf{0.245} \\
        \hline
        \hline
        \textbf{Dataset} &
        \multicolumn{3}{c|}{\textbf{Video Games}} &
        \multicolumn{3}{c|}{\textbf{Pet Supplies}} &
        \multicolumn{3}{c|}{\textbf{Office Products}}  \\
        \hline
        \textbf{Metrics} & 
        Hit@3 & NDCG@3 & MRR &
        Hit@3 & NDCG@3 & MRR &
        Hit@3 & NDCG@3 & MRR \\
        \hline
        \textbf{CF} & 0.025 & 0.014 & 0.030 & 0.030 & 0.017 & 0.031 & 0.036 & 0.021 & 0.030\\
        \textbf{MF} & 0.020 & 0.015 & 0.022& 0.018 & 0.014 & 0.023 & 0.008 & 0.006 & 0.010\\
        \textbf{Knowledge-Aware} & 0.021 & 0.013 & 0.028& 0.017 & 0.011 & 0.022 & 0.010 & 0.007 & 0.013\\
        \textbf{LightGCN} & {\tt OOME} & {\tt OOME} & {\tt OOME}& {\tt OOME} & {\tt OOME} & {\tt OOME} & {\tt OOME} & {\tt OOME} & {\tt OOME}\\ \hline
        \textbf{\name$_{\mathrm{Low}}$} & \textbf{0.095} & \textbf{0.071} & \textbf{0.104} & 0.061 & 0.044 & 0.066& 0.034 & 0.025 & 0.037\\
        \textbf{\name$_{\mathrm{High}}$} & 0.001 & 0.001 & 0.001 & 0.001 & 0.001& 0.001 & 0.001 & 0.001 & 0.001\\
        \textbf{\name$_{\mathrm{Hybrid}}$} & 0.079 & 0.057 & 0.090& \textbf{0.133} & \textbf{0.097} & \textbf{0.137}& \textbf{0.039} & \textbf{0.029} & \textbf{0.042} \\
        \hline
    \end{tabular}
    \vspace{2.5mm}
\end{table*}

\begin{table*}[!t]
\captionsetup{font=small}
    \centering
    \caption{Evaluation results for high-order complementary recommendation. ({\tt OOME} stands for out-of-memory error)} 
    \label{tab:high-order}
    \scalebox{1}{
    \begin{tabular}{|c|ccc|ccc|ccc|}
        \hline
        \textbf{Dataset} &
        \multicolumn{3}{c|}{\textbf{Gift Cards}} &
        \multicolumn{3}{c|}{\textbf{Digital Music}} &
        \multicolumn{3}{c|}{\textbf{Amazon Fashion}} \\
        \hline
        \textbf{Metrics} & 
        Hit@3 & NDCG@3 & MRR &
        Hit@3 & NDCG@3 & MRR &
        Hit@3 & NDCG@3 & MRR \\
        \hline
        \textbf{CF} & 0.029 & 0.015 & 0.052& 0.086 & 0.047 & 0.088& 0.080 & 0.042 & 0.058\\
        \textbf{MF} & 0.044 & 0.025 & 0.065& 0.013 & 0.007 & 0.012& 0.019 & 0.014 & 0.018\\
        \textbf{Knowledge-Aware} & 0.042 & 0.024 & 0.061& 0.097 & 0.067 & 0.091 & 0.027 & 0.020 & 0.031\\
        \textbf{LightGCN} & 0.054 & 0.037 & 0.079& 0.067 & 0.046 & 0.053& {\tt OOME} & {\tt OOME} & {\tt OOME}\\\hline
        \textbf{\name$_{\mathrm{Low}}$} & 0.060 & 0.045 & 0.068& 0.335 & 0.260 & 0.291& 0.047 & 0.035 & 0.047\\
        \textbf{\name$_{\mathrm{High}}$} & 0.192 & 0.136 & 0.197& 0.566 & 0.414 & 0.432& \textbf{0.520} & \textbf{0.369} & \textbf{0.397}\\
        \textbf{\name$_{\mathrm{Hybrid}}$} & \textbf{0.256} & \textbf{0.184} & \textbf{0.241}& \textbf{0.641} & \textbf{0.485} & \textbf{0.495}& 0.470 & 0.353 & 0.387 \\
        \hline
         \hline
        \textbf{Dataset} &
        \multicolumn{3}{c|}{\textbf{Video Games}} &
        \multicolumn{3}{c|}{\textbf{Pet Supplies}} &
        \multicolumn{3}{c|}{\textbf{Office Products}} \\
        \hline
        \textbf{Metrics} & 
        Hit@3 & NDCG@3 & MRR &
        Hit@3 & NDCG@3 & MRR &
        Hit@3 & NDCG@3 & MRR \\
        \hline
        \textbf{CF} & 0.030 & 0.015 & 0.044 & 0.018 & 0.009 & 0.023& 0.032 & 0.017 & 0.035\\
        \textbf{MF} & 0.015 & 0.011 & 0.016 & 0.003 & 0.002 & 0.004& 0.005 & 0.004 & 0.005\\
        \textbf{Knowledge-Aware} & 0.044 & 0.029 & 0.057& 0.014 & 0.010 & 0.019 & 0.015 & 0.011 & 0.018\\
        \textbf{LightGCN} & {\tt OOME} & {\tt OOME} & {\tt OOME}& {\tt OOME} & {\tt OOME} & {\tt OOME}& {\tt OOME} & {\tt OOME} & {\tt OOME}\\\hline
        \textbf{\name$_{\mathrm{Low}}$} & 0.043 & 0.031 & 0.053& 0.007 & 0.005 & 0.010& 0.001 & 0.001 & 0.002\\
        \textbf{\name$_{\mathrm{High}}$} & \textbf{0.144} & 0.102 & \textbf{0.152}& 0.060 & 0.044 & 0.070& 0.009 & 0.006 & 0.015\\
        \textbf{\name$_{\mathrm{Hybrid}}$} & \textbf{0.144} & \textbf{0.103} & 0.151 & \textbf{0.099} & \textbf{0.072} & \textbf{0.103}& \textbf{0.038} & \textbf{0.027} & \textbf{0.051} \\
        \hline
    \end{tabular}
    }
\end{table*}

\vspace{-1mm}
\subsection{Experiment Setup}

\textbf{Dataset:} We utilize the Amazon dataset~\cite{ni2019justifying}, which was collected as a user-item-review graph. We conduct experiments on its top-level product categories\footnote{\url{https://jmcauley.ucsd.edu/data/amazon/}}: Gift Cards, Digital Music, Amazon Fashion, Video Games, Pet Supplies, and Office Products. 
The statistics of the six datasets are summarized in TABLE~\ref{tab:dataset}. Specifically, we adopt two types of relationships for model learning, \ie, the co\_purchase, where users who bought item $i$ also bought item $j$; and the co\_view, where users who viewed item $i$ also viewed item $j$. We use triplets $(h, r, t)$ to store our data, where $h$ is the head entity, $t$ is the tail entity, and $r$ is the relationship. Particularly, for low-order data, we can directly extract triplets from the Amazon dataset, where $h$ and $t$ are items, and $r$ is the relationship (namely, co\_purchase and co\_view); for high-order data, $h$ is a tuple of two items that are complementary with each other, $t$ is another item which is complementary to both items in the head, and co\_purchase is the only relationship $r$ here. We generate high-order data from low-order data based on the following rule: 
if $i_{1} \rightarrow i_{3}$, $i_{2} \rightarrow i_{3}$, and $i_{2} \rightarrow i_{1}$ (or $i_{1} \rightarrow i_{2}$), then we derive the high-order complementary pair ($i_{1}, i_{2}) \rightarrow i_{3}$.

\textbf{Baselines:} We conduct comparison experiments between \name\ and the following recommendation baselines:
\begin{desclist}
\item {\bfseries Collaborative Filtering} (CF)~\cite{linden2003amazon}: It is the most classical method where each item is represented as its corresponding multi-hot user interaction vector. The complementary relationship between items is captured by their vector product; 
\item {\bfseries Matrix Factorization} (MF)~\cite{koren2009matrix}: The method learns the item representation by decomposing the user-item interaction matrix, and thus conducts complementary recommendation; 
\item{\bfseries LightGCN}~\cite{he2020lightgcn}: User and item embeddings are learned on the graph. Final item representation is the average of corresponding user embeddings that have interactions with the current item; 
\item {\bfseries Knowledge-Aware}~\cite{xu2020knowledge}: Dual embedding is learned to represent items from contextual knowledge by multi-task learning. We consider (1) the most recent five ratings from reviews as the contextual information, (2) the sequence in which items are reviewed as the order they are purchased.
\item {\bfseries \name\ variations}: We conduct ablation studies by comparing \name\ with its variations, including \name$_{\mathrm{Low}}$ (the blue region of Figure~\ref{fig:framework}) which is solely trained based on the low-order data; \name$_{\mathrm{High}}$ (the orange region of Figure~\ref{fig:framework}) which is solely trained based on the high-order data; and \name$_{\mathrm{Hybrid}}$ which is trained based on both low-order data and high-order data.
\end{desclist}

\noindent\textbf{Evaluation Metrics:} We randomly split the dataset into 70\%, 20\%, 10\% for model training, validation, and testing. 
We evaluate the performance of all methods by three standard measurements for ranking tasks:  Hit Rate, Normalized Discounted Cumulative Gain (NDCG), and Mean Reciprocal Rank (MRR).
In these experiments, Hit@3, NDCG@3, and MRR are used to report the performance.

\noindent\textbf{Implementation Details:}
The datasets are publicly available, and the code has been released on GitHub\footnote{\url{https://github.com/wulongfeng/LogiRec}}. For all the reported results, we set $m$ = 128, $\gamma$ = 60, $k$ = 400. The experiments are mainly performed on a Linux system with an NVIDIA GeForce RTX 3090 graphics card (24GB GDDR6X).

\begin{figure}[h]
\small
\captionsetup{font=footnotesize}
\centering
\subfigure[Hit@3]{
    \includegraphics[width=2.9cm, height=3.5cm]{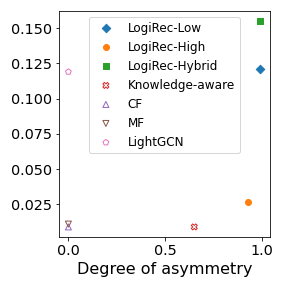}
    }
\hspace{-4.5mm}
\subfigure[NDCG@3]{
    \includegraphics[width=2.9cm, height=3.5cm]{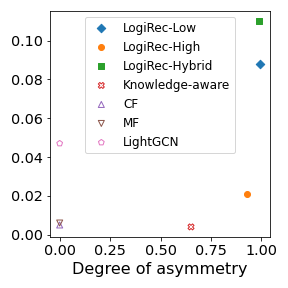}
    }
\hspace{-4.8mm}
\subfigure[MRR]{
    \includegraphics[width=2.9cm, height=3.5cm]{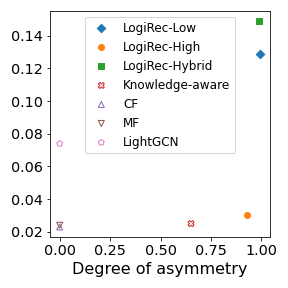}
    }
\label{fig:label13} 
\caption{Degree of asymmetry on Gift Cards dataset.}
\label{fig:degree of asymmetry} 
\vspace{-2mm}
\end{figure}
\vspace{1.0mm}

\subsection{Ranking Comparison}
We evaluate the effectiveness of \name\ and its variants on both low-order CR and high-order CR settings by comparing with four baseline methods across three evaluation metrics. The results are reported in TABLE~\ref{tab:low-order} and TABLE~\ref{tab:high-order}. Please note that, due to the out-of-memory error, we do not list the performance of LightGCN on the dataset of Amazon Fashion, Video Games, Pet Supplies, and Office Products. 
In general, we have the following observations: 
(1) \name\ and its variants outperform most of the baseline methods with a large margin across all the metrics and datasets. For instance, in Gift Cards, \name\ achieves 1.7\%, 89\% and 79\%  higher performance regarding Hit@3, NDCG@3 and MRR than our best competitor LightGCN.
(2) \name$_{\mathrm{Hybrid}}$ achieves the best performance on most metrics. \name$_{\mathrm{Low}}$ has competitive results on the low-order CR setting, so does \name$_{\mathrm{High}}$ on the high-order CR setting. We believe this is because the model's architecture is well designed with dataset, and \name$_{\mathrm{Hybrid}}$ effectively integrates the information of \name$_{\mathrm{High}}$ and \name$_{\mathrm{Low}}$.
(3) The performance of CF is better than MF, we believe this is due to two factors: MF is better at capturing information between substitute items; Amazon's recommendation solution is based on CF~\cite{linden2003amazon}, thus CF could benefit from the final presentation of items.

\subsection{Asymmetric Comparison}
To demonstrate the asymmetric property of \name, we conduct a comparison among \name\ and the baseline methods on Gift Cards dataset, which is shown in Figure~\ref{fig:degree of asymmetry}. 
The x-axis shows the degree of asymmetry, and y-axis shows the performance on the low-order CR setting (Hit@3, NDCG@3 and MRR).
The degree of asymmetry is defined as follows.
\begin{equation} 
\small
\begin{split}
asy = \frac{\sum_{p=1}^{|\mathcal{I}|}\sum_{q=p+1}^{|\mathcal{I}|}|P(i_{p}|i_{q}) - P(i_{q}|i_{p})|}{|\mathcal{I}|}
\end{split}
\end{equation}
where $|\mathcal{I}|$ is the number of nodes in the dataset. 
The method with better performance would be located in the upper region of figures, with higher asymmetry would be located in the right region.
From Figure~\ref{fig:degree of asymmetry}, we could conclude that: (1) CF, MF and LightGCN are symmetric, and knowledge-aware and our methods are asymmetric, which is consistent with our understanding. (2) Our methods are located in the upper right region, indicating that our methods are asymmetric and perform better than the baseline methods.

\section{Related Work}
\textbf{Recommendation systems.} With the information explosion of the big data era, recommendation systems play a pivotal role in alleviating information overload. Various approaches have been developed, the CF-based methods are the most mature and widely used in the real-world systems. 
Recently, Graph Convolutional Networks (GCNs)~\cite{he2020lightgcn} have also been proven to be effective on recommendation tasks due to their expressive power in modeling both user-item interactions and the graph structures.
However, the representation of entities and relationships learned by these methods are all symmetric. 
To accommodate the asymmetric nature of complementary recommendation, some asymmetric representation learning approaches have been developed. Two typical examples are P-Companion~\cite{hao2020p} and knowledge-aware complementary product representation~\cite{xu2020knowledge}. P-Companion jointly predicts the complementary products and the complementary product types, and knowledge-aware utilizes dual embedding to represent products.
However, these methods are all designed for low-order CR, thus fail to explore the high-order CR setting.

\textbf{Logical reasoning on graphs.} With the success of representation learning, graph embedding methods have also gained successes via learning the heterogeneous latent representations of product entities. Meanwhile, logical reasoning on graphs has received a surge of research interest from the data mining community.
Early logical reasoning models mine and discover new facts based on logical rules or statistical features~\cite{schoenmackers2010learning}, which rely heavily on the logical rules. 
Due to the complexity and diversity of entities and relationships in large-scale knowledge graphs, it is difficult to exhaust all inference patterns. 
In the past decade, embedding-based approaches have gained growing attention~\cite{bordes2013translating,sun2019rotate}, most of which support relationship projection. Some methods extend the idea of projection and define a query language that supports intersection~\cite{hamilton2018embedding}, filtering~\cite{sun2020faithful}, and difference~\cite{liu2021neural}. BetaE~\cite{ren2020beta} directly gives the definition of negation operation which could handle arbitrary first-order logical operations.
In this paper, we utilize a BetaE-based model to define the operation of complementary recommendation and extend it to high-order scenarios.

\section{Conclusion}
Exploring complementary relationships plays a pivotal role in the modern recommendation systems. Despite the key importance, existing work is mostly designed for the low-order complementary recommendation settings while failing to model the high-order dependencies among products. In this paper, we present an end-to-end framework named \name, which automatically captures the asymmetric complementary relationship between products and seamlessly extends to the high-order complementary recommendation setting. Extensive results show that the proposed \name\ framework achieves significant improvements in the settings of the low-order complementary recommendation and the high-order complementary recommendation by comparing with three popular baseline methods across three evaluation metrics.

\bibliographystyle{IEEEtranN}
\footnotesize
\bibliography{reference_short}

\vspace{12pt}

\end{document}